\documentclass[twoside,11pt]{article}

\usepackage{blindtext}

\usepackage{jmlr2e}
\usepackage{amsmath}
\usepackage{algorithm}
\usepackage{algorithmic}

\newcommand{\dataset}{{\cal D}}
\newcommand{\fracpartial}[2]{\frac{\partial #1}{\partial  #2}}

\usepackage{lastpage}
\ShortHeadings{Enhancing Carbon Emission Reduction Strategies}{Åström, Geldhauser, Grillitsch, Hall, Sopasakis}
\firstpageno{1}

\begin{document}

\title{Enhancing Carbon Emission Reduction Strategies using OCO and ICOS data}

\author{\name Oskar Åström \email oskar.astrom@math.lth.se \\
     \name Alexandros Sopasakis \email alexandros.sopasakis@math.lth.se \\
   \addr  Centre for Mathematical Sciences \\
    Lund University \\
    221 00 Lund, Sweden\\
    \name Carina Geldhauser \email carina.geldhauser@ma.tum.de\\
       \addr Munich Center for Machine Learning \\
       Technical University of Munich \\
       85748 Garching, Germany \\
   \name Markus Grillitsch \email markus.grillitsch@keg.lu.se \\
     \name Ola Hall \email  ola.hall@keg.lu.se\\
     \addr Department of Human Geography \\
Lund University \\
221 00 Lund, Sweden
       }

\editor{TBD}

\maketitle

\begin{abstract}%   <- trailing '%' for backward compatibility of .sty file
We propose a methodology to enhance local CO\textsubscript{2} monitoring by integrating satellite data from the Orbiting Carbon Observatories (OCO-2 and OCO-3) with ground level observations from the Integrated Carbon Observation System (ICOS) and weather data from the ECMWF Reanalysis v5 (ERA5). Unlike traditional methods that downsample national data, our approach uses multimodal data fusion for high-resolution CO\textsubscript{2} estimations.

We employ weighted K-nearest neighbor (KNN) interpolation with machine learning models to predict ground level CO\textsubscript{2} from satellite measurements, achieving a Root Mean Squared Error of 3.92 ppm. Our results show the effectiveness of integrating diverse data sources in capturing local emission patterns, highlighting the value of high-resolution atmospheric transport models. The developed model improves the granularity of CO\textsubscript{2} monitoring, providing precise insights for targeted carbon mitigation strategies, and represents a novel application of neural networks and KNN in environmental monitoring, adaptable to various regions and temporal scales.
\end{abstract}

\begin{keywords}
Carbon Emission Monitoring, ICOS, Machine Learning, OCO-2 and OCO-3, Remote Sensing, Ground Level CO\textsubscript{2}
 % keyword one, keyword two, keyword three
\end{keywords}

\newpage
\section{Introduction\label{introduction}}
In this era of rapid climate change, the precise monitoring of carbon dioxide emissions has become a critical task for governing bodies and environmental organizations worldwide. The ability to accurately assess and manage CO\textsubscript{2} levels is not just a scientific challenge, but a necessity for effective climate action.

%This manuscript delves into utilizing satellite data to extrapolate municipality-level CO\textsubscript{2} budgets. The proposed methodology  therefore explores a leap from traditional reliance on national data downsampling or extrapolation from sparse ground measurements, which often hinders the effectiveness of climate action monitoring and the assessment of mitigation strategies.

The current landscape of carbon dioxide~{(CO\textsubscript{2})} measurement is fraught with challenges, predominantly due to the lack of comprehensive and localized data. Local authorities, which are on the front line of implementing climate action strategies, often find themselves relying on national-level data or extrapolations from limited ground measurements. This lack of granularity in the data can lead to inaccuracies in identifying local emission sources, while also impeding the development of more effective, targeted, climate strategies. 

Our research aims to bridge this gap by using satellite-derived data to provide a more precise and localized view of CO\textsubscript{2} emissions.
We adopt a multimodal data approach, %that encompasses a wide range of data collection and analysis methods. Central to this approach is the utilization of 
utilizing data from the Orbiting Carbon Observatories (OCO-2 and OCO-3) as well as the Integrated Carbon Observation System (ICOS) and the ECMWF Reanalysis v5 (ERA5) model, enabling us to establish vital connections between satellite transects, ground level CO\textsubscript{2} data, and weather. 
%Specifically, data was facilitated through different platforms:
%\begin{itemize}
%    \item 32 ICOS stations across Europe as ground truth data since 2017. The data was accessed via an API provided by ICOS for Python;
%    \item OCO-2 and OCO-3 Level 2 data, whole mission (July 2014 - March 2023),  was obtained through NASA's OPeNDAP service;
%    \item Bias corrected, ERA5 weather data was retrieved via the Climate Data Store API with MARS requests.
%\end{itemize}
This integration is pivotal in enhancing the precision and reliability of CO\textsubscript{2} datasets derived from satellite observations, offering an unprecedented level of detail in emission monitoring.

In technical terms, our aim is to build a machine learning algorithm which,  using satellite data for carbon dioxide, weather, and other relevant input factors, can model accurate measurements, similar to ICOS ground data measurements of the biospheric surface CO\textsubscript{2} flux anomalies.

However, monitoring of CO\textsubscript{2} using satellite data is not without its challenges. Unlike pollutants with relatively short atmospheric lifetimes such as NO$_2$ and SO$_2$,  CO\textsubscript{2} has a longer lifespan, resulting in a high background concentration that complicates the identification of new anthropogenic sources. Furthermore, the spatiotemporal distribution of CO\textsubscript{2} is heavily influenced by atmospheric dynamics and terrestrial biospheric fluxes, introducing significant variability and seasonality in measurements.

Although significant strides have been made in atmospheric CO\textsubscript{2} monitoring, the complexity of the task remains, with challenges in regions of complex topography and varying meteorological conditions. Continuous advancements in satellite technology, data processing, and atmospheric modeling are crucial to enhance the accuracy and scope of global CO\textsubscript{2} monitoring.

Our methodology addresses these challenges by creating machine learning models that employ advanced statistical analyses to estimate ground observations of CO\textsubscript{2} from satellite data. We also explore the utility of high-resolution satellite data in capturing small-scale variability of CO\textsubscript{2}, which is often lost in coarser-resolution models. This approach is critical in understanding the nuances of CO\textsubscript{2} weather - the interaction between weather patterns and CO\textsubscript{2} surface fluxes – and in providing accurate data for regional modeling studies and field experiments.
We begin in Section \ref{RW} with an overview of satellite-based CO\textsubscript{2} monitoring approaches and their utilization in CO\textsubscript{2} estimation.

 Then, in Section \ref{OCO} we describe current methodological advances as well as challenges in the data collected from the OCO-2 and OCO-3 satellites. We then describe in more detail the methods and data used in this work in Section \ref{Method}. We present our findings and end with a feature importance analysis of the models in Section \ref{results}.

\section{State of the art\label{RW}}
 
In this section, we provide an overview of key satellite instruments that have significantly advanced the field of CO\textsubscript{2} monitoring. We discuss the specific contributions of satellites such as SCIAMACHY, GOSAT, and OCO-2, highlighting their roles in improving the precision of CO\textsubscript{2} measurements and their applications in environmental research. The discussion includes an analysis of how these satellites have been utilized to detect CO\textsubscript{2} emissions from various sources, understand the effects of atmospheric transport on CO\textsubscript{2} distribution, and enhance our overall understanding of the global carbon cycle.

In recent decades, significant advances have been made in satellite technology for environmental monitoring. Notable among these are instruments like SCIAMACHY (Scanning Imaging Absorption Spectrometer for Atmospheric Chartography) onboard the Environmental Satellite, ENVISAT, operational from 2002 to 2012 and the Greenhouse Gases Observing Satellite (GOSAT), launched in 2009. These satellites, along with others such as the American OCO-2, the Chinese TanSat, and the upcoming Geostationary Carbon Observatory (GeoCarb), have revolutionized our ability to monitor atmospheric CO\textsubscript{2}. Each of these satellites offers unique contributions to the field, from high-precision CO\textsubscript{2} observing capabilities to the monitoring of CO\textsubscript{2} variations on seasonal time scales.

The analysis of data collected by SCIAMACHY has been instrumental in deriving empirical regional conversion factors, which are used to estimate CO\textsubscript{2} emissions based on observed NO\textsubscript{2} columns. This approach, as demonstrated in previous studies \cite{schneising2013remote, reuter2014remote}, helps establish a crucial link between NO\textsubscript{2}—a short-lived pollutant indicative of fossil fuel combustion—and the longer-lived CO\textsubscript{2}, thereby enhancing our understanding of anthropogenic emissions and their impact on atmospheric composition.

SCIAMACHY data was used also to estimate CO\textsubscript{2} abundance from fossil fuel emissions, particularly from power plants \cite{Bovensmann2010}. This approach has provided insights into the spatial distribution of CO\textsubscript{2} emissions, contributing to a more nuanced understanding of anthropogenic impacts on atmospheric CO\textsubscript{2} levels.

%GOSAT
Greenhouse gases observing satellite (GOSAT) observations have been pivotal in detecting CO\textsubscript{2} emission signatures from urban centers like Los Angeles and Mumbai \cite{kort2012four, yokota2009global}. This detection capability is crucial for understanding the role of megacities in global CO\textsubscript{2} emissions. These satellite-based CO\textsubscript{2} abundances can be compared with emission inventories, revealing potential gaps in emission estimates, as was shown in a study from East Asia \cite{janardanan2016comparing}.

The capabilities of OCO-2 surpass those of earlier instruments such as SCIAMACHY and GOSAT, particularly due to a significant reduction in xCO\textsubscript{2} retrieval uncertainty. This reduction is achieved through OCO-2's higher spectral resolution, advanced data processing algorithms, and improved spatial and temporal coverage. As a result, OCO-2 provides more precise and reliable measurements of CO\textsubscript{2}, enhancing our ability to monitor global emissions, validate emission inventories, and improve climate models, thereby deepening our understanding of CO\textsubscript{2}'s role in climate change.

Feldman et al. \cite{Feldman2023} showed that OCO-2 can also detect anomalies arising from terrestrial biosphere extremes, for example droughts and heat waves with a rate of 80\% in some parts of Northern Australia and 'greater than by chance' detection rates for the most extreme CO\textsubscript{2} surface flux anomalies in western US (\cite{Feldman2023}, page 1555). This advancement is crucial for detecting subtle CO\textsubscript{2} anomalies that may be indicative of significant environmental events or trends.

\subsection{Challenges in CO\texorpdfstring{\textsubscript{2}}{2} emissions estimation}

Carbon dioxide presents unique challenges due to its longer atmospheric lifespan and high background concentration, which complicates the identification of anthropogenic sources \cite{Hakkarainen2023, Bovensmann2010, Keppel-Aleks2013}.
%
% Bottom-up approaches
The bottom-up compilation of CO\textsubscript{2} inventories, based on reported data or human activity such as energy consumption and fuel purity, faces notable discrepancies in emission estimates.  These uncertainties are especially pronounced in megacities in developing countries \cite{peylin2013, wang2013}.

Due to its longer lifespan, atmospheric transport of CO\textsubscript{2} becomes a relevant factor, which can lead to overestimation in deserts and underestimation in areas with significant vegetation, as indicated by He et al. \cite{he2018satellite}. Their study highlights how extreme changes in atmospheric CO\textsubscript{2} concentrations, detected using satellite data, are influenced by local environmental conditions and transport dynamics, underscoring the complexity of accurately estimating CO\textsubscript{2} levels in diverse regions.

The variability of atmospheric CO\textsubscript{2} is heavily influenced by its interaction with weather patterns, which requires high-resolution atmospheric transport models. These models are critical for capturing small-scale variability and understanding the intricacies of CO\textsubscript{2} weather \cite{Feldman2023}. However, monitoring global carbon sources and sinks is highly complex and needs to quantify longer latency and errors due to assumptions about uncertain surface CO\textsubscript{2} flux drivers and meteorological conditions \cite{Feldman2023}.

% Weather data
The integration of weather data, as explored by Hakkarainen et al. \cite{Hakkarainen2023}, Beirle et al. \cite{Beirle2011}, and Fioletov et al. \cite{fioletov2015lifetimes}, offers a more comprehensive understanding of emissions, particularly from isolated urban areas. Incorporating wind speed information is crucial for interpreting emission patterns and understanding the dynamics of pollutant dispersion. Research indicates the need for undisturbed atmospheric transport for accurate xCO\textsubscript{2} retrievals, highlighting the need for low variability in wind direction and speed, a consistent wind source and the avoidance of complex topography near the surface \cite{Hakkarainen2023, Beirle2011, fioletov2015lifetimes, Feldman2023}.

\subsection{OCO-2 and OCO-3: Advances and Challenges in CO\texorpdfstring{\textsubscript{2}}{2} Monitoring}\label{OCO}

The Orbiting Carbon Observatory satellites, OCO-2 and OCO-3, represent a significant advancement in the global monitoring of carbon dioxide (CO\textsubscript{2}). Unlike direct measurement techniques, these satellites detect CO\textsubscript{2} by measuring the absorption of sunlight reflected from the Earth's surface within an air column \cite{Eldering2017, Taylor2023}. This approach, combined with their high sensitivity, enables these satellites to provide detailed insights into atmospheric CO\textsubscript{2} concentrations. The measurement area for OCO-2 and OCO-3 is approximately $1.29 \times  2.25$ km, achieved through a sun-synchronous orbit that ensures consistent data acquisition across different regions \cite{Feldman2023}.

A key strength of OCO-2 and OCO-3 lies in their ability to generate column-averaged dry-air mole fractions of CO\textsubscript{2}. This is accomplished through advanced retrieval algorithms, which integrate point measurements to produce highly accurate data for further analysis \cite{OCO2DataCenter}. The precision of these satellites has been instrumental in urban emission monitoring, particularly in evaluating fossil fuel CO\textsubscript{2} emissions from densely populated areas. High-resolution transport modeling, combined with Bayesian inversion systems, has been used to optimize city-wide emission estimates \cite{Ye2020}. Furthermore, the Snapshot Area Mapping (SAM) mode of OCO-3 has proven effective in characterizing major anthropogenic sources, especially when supplemented with NO$_2$ measurements from other satellite instruments \cite{Hakkarainen2023}.

However, despite these advancements, OCO-2 and OCO-3 face several challenges that limit their effectiveness in certain scenarios. Firstly, their sun-synchronous orbits and narrow measurement swaths restrict the frequency and coverage of observations over specific regions, especially concerning point sources such as power plants, where emission origins are concentrated to a single location.

Additionally, both satellites are subject to retrieval errors and observation gaps, which can hinder the consistent detection of flux anomalies, particularly in complex terrains or under specific atmospheric conditions \cite{Taylor2023}. The accuracy of data can also be affected by biases introduced through small pointing errors, especially in regions with rough topography. Therefore, rigorous data filtering and bias correction, calibrated against standards like TCCON and the WMO CO\textsubscript{2} reference scale, are crucial to maintaining data integrity \cite{OCO2DataCenter}.

The modeling of point sources, such as coal power plants \cite{Feldman2023}, further highlights the challenges  associated with satellite-based CO\textsubscript{2} monitoring. For instance, several studies \cite{ODell2018, Claeyman2011, Wang2022} have shown that a significant proportion of satellite data can be rendered unusable due to unfavorable atmospheric conditions, such as cloud cover, aerosols, or haze, which obscure the satellite's view and lead to retrieval errors. Even when data retrieval is successful, such measurements may fail to capture information near or even downwind of the emission source due to, for instance, sudden changes in wind speed and misalignment between the emission source and the detected plume \cite{Feldman2023}. These factors underscore the need for continuous refinement of satellite-based CO\textsubscript{2} monitoring techniques in order to improve data reliability and accuracy.

While OCO-2 and OCO-3 have significantly advanced our capability to monitor global CO\textsubscript{2} emissions, ongoing challenges in data retrieval, particularly over complex terrains and smaller emission sources, underscore the need for continued refinement in both satellite technology and data processing methodologies.

\section{Materials and Methods}\label{Method}
%\subsection{Data Description}\label{Data Description}

This study uses data from multiple openly available high-resolution datasets, providing comprehensive measurements of the CO\textsubscript{2} concentration and the atmospheric variable, which we describe in the following section.

\subsection{Datasets}

We began by employing OCO-2 and OCO-3 satellite data consisting of bias-corrected, retrospective xCO\textsubscript{2} measurements from the Level 2 mission\footnote{Available at \url{https://disc.gsfc.nasa.gov/datasets/OCO2_L2_Lite_FP_10r/summary}}. To ensure accuracy and reliability, this data is pre-processed at the source by the OCO-2 and OCO-3 Science Teams at NASA and associated institutions \cite{OCO2_L2_Lite_FP_10r}. This pre-processing also includes physics-based algorithms to adjust for systematic errors through bias correction and produce improved measurements of the mole fraction of CO\textsubscript{2} in dry air, as determined from the sun absorption spectra in the near-infrared at 1.61 and 2.06 $\mu$m.
The satellite data, characterized by a spatial resolution within a one-hour time frame, covers a parallelogram-shaped area of 2.29 km\(^2\) per column with an orbit track width of 10.3 km. 
Due to the slow nature of API requests, the retrieval of weather data was performed on a one-degree latitude and longitude grid and restricted to a six-hour window from 9:00 to 15:00.

 Second, we utilized ICOS station data, incorporating measurements from 32 ICOS atmospheric stations across Europe since 2017. These stations provide hourly averages of dry-air mole fractions of CO\textsubscript{2}. They employ nondispersive infrared (NDIR) analyzers for the spectral analysis of air samples. Unlike the OCO dataset, ICOS data are geographically fixed and offer continuous time series data.
However, it is important to note that the recordings are not always complete due to various factors such as instrument malfunctions, power failures, extreme weather conditions, and regular maintenance, which result in interrupted data at most stations.

 Third, we used ERA5 Reanalysis Weather Data, sourced from the Copernicus programme and the ECMWF's Climate Data Store. This dataset provided supplemental weather data, including wind components at $10m$ height, surface-level pressure, and temperature at $2m$. Additional variables such as the vertical integral of temperature and total column water vapor, cloud-based height, and total cloud cover were also included.

%\subsection{Temporal Coverage and Geographical Scope}
The study period spanned from the beginning of the OCO mission in September 2014 to March 2023. The spatial coverage for the weather data for OCO, ICOS and ERA5 was defined within a geodesic rectangle bounded by coordinates $28^\circ S$, $-17^\circ W$, $70^\circ N$ and $64^\circ E$. The sun-synchronous orbit of the OCO satellite resulted in more frequent data collection over European ICOS stations during summer and reduced coverage in winter. 

\begin{figure}[ht]
    \centering
    \includegraphics[width=.9\linewidth]{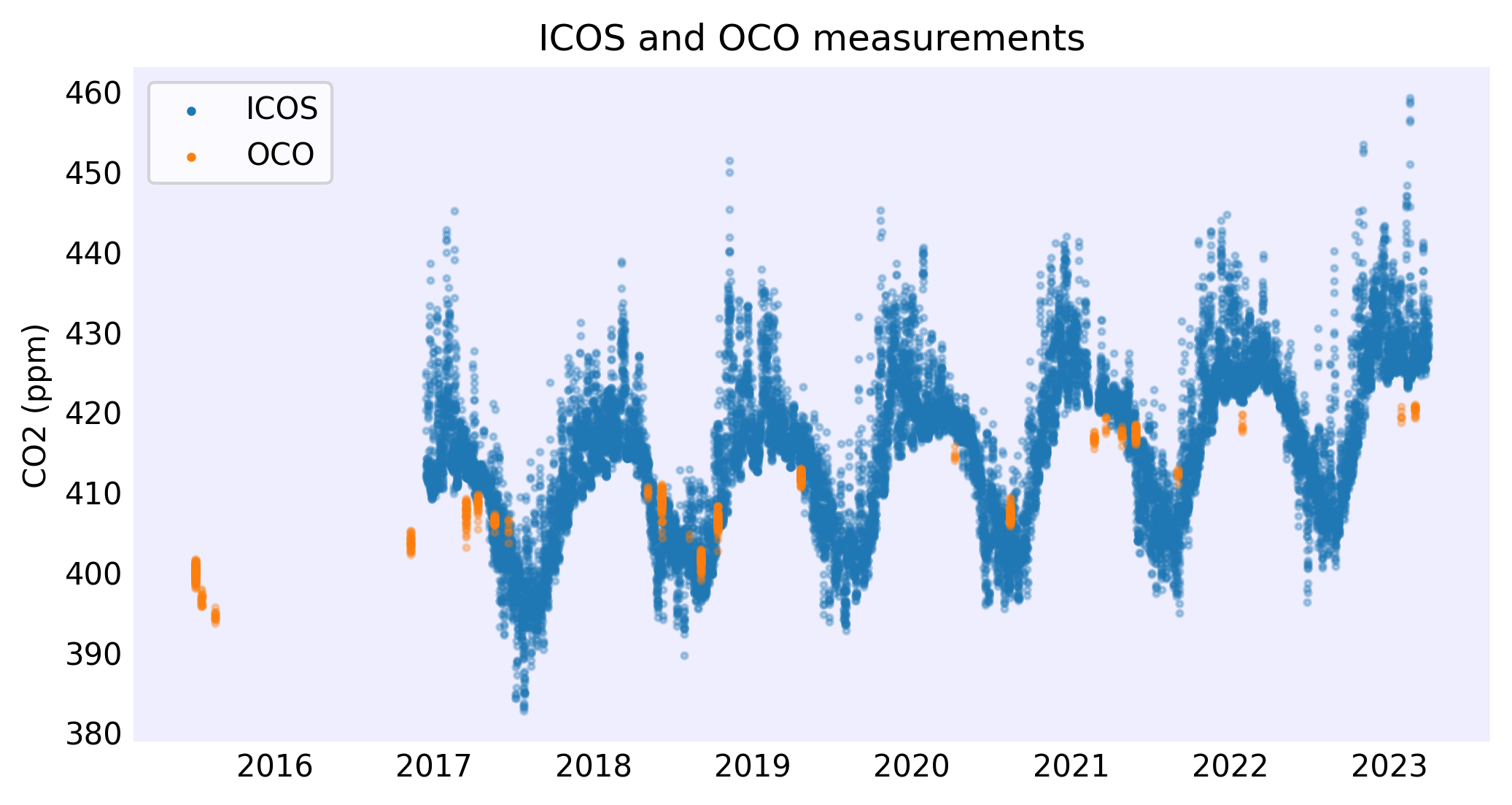}
    \caption{ Measurements from the Hyltemossa ICOS station (blue) and nearby OCO observations (orange) within a 25km radius for the time frame of data availability.  }
    \label{fig:measurements_time_series}
  \end{figure}

To construct the dataset, satellite measurements from OCO was matched with the closest measurement from a station, if it was within a certain distance in space and time. While the ICOS stations measured continuously, data could only be combined when the satellite passed over which reduced the amount of data drastically. Figure \ref{fig:measurements_time_series} shows the ICOS and OCO measurements for a region around the Hyltemossa ICOS station. The figure shows the small amount of times where there is a corresponding OCO measurement near the station. Figure \ref{fig:measurements_time_series} also shows that the OCO measurements are highly correlated with the ICOS data, as it seems to follow a similar one-year cycle in the CO\textsubscript{2} levels. 

%\subsubsection{Data Augmentation and Preprocessing}
To increase the variance in the input data, the OCO measurements were augmented with ERA5 weather data and the location and time information of the measurement.
ICOS stations record weather in addition to CO\textsubscript{2}. However, since these stations are geographically fixed and primarily record wind speed and direction, using ICOS data as input prohibits the use of this model outside of the station bounds. 
Consequently, ERA5 data was used for a more comprehensive weather analysis as it has global reach and records more weather features.

\subsection{Machine Learning Models\label{sec:Models}}

To estimate CO\textsubscript{2} levels from satellite data, we used multiple classes of machine learning models. 

As a baseline, we perform simple linear regression on ICOS measurements based only on OCO observations. This baseline therefore does not contain any weather data and is a measurement of how well the OCO satellite data itself correlates with the ground level CO\textsubscript{2}.

Secondly, we employ Boosting models, chosen due to their proven efficacy in handling complex non-linear data relationships, which is characteristic of environmental datasets \cite{wen2021features, cai2020prediction}. Boosting models
in general are a type of ensemble machine learning method designed to improve the accuracy of predictions by combining the outputs of multiple weak learners into a single strong model. The key idea behind boosting is to sequentially train weak models, where each model focuses on correcting the errors made by the previous models in the sequence. 
The ensemble of weak models therefore is improved by iteratively adding new models that are trained to predict the pseudo-residuals of the previous iteration of the ensemble. These models have been shown to outperform classical methods like Stochastic Gradient Descent and Support Vector Machines in environmental monitoring \cite{cai2020prediction}. 

We employed two different types of boosting models; Category Boosting and Extreme Gradient Boosting. Extreme Gradient Boosting is a boosting model that uses continuous input features to describe the data, whereas Category Boosting employs categorical features and outputs. This discretizes the data into less descriptive but more complex classes, rather than a point in N-dimensional space. The parameters and packages used for Category and Extreme Gradient Boosting are specified in the Appendix.

In addition to employing Boosting models, we utilized a multilayer perceptron (MLP) neural network tailored for regression tasks. This MLP comprises four densely connected hidden layers, encompassing a total of 19,649 trainable parameters—a relatively modest architecture by contemporary standards. This compact model was deliberately chosen due to the nature of the dataset, which includes only 14 features that exhibit strong correlations with the ground truth. Furthermore, since the measurements were exclusively obtained from locations surrounding the ICOS stations, a more complex model would likely lead to significant overfitting to the training data. Further details on the model's architecture and implementation can be found in the Appendix.

Each model was trained on an identical training set comprising 31,039 data points and evaluated on a test set of 7,584 data points. The test data was sourced from four ICOS stations that were intentionally excluded during training. This strategy was chosen to prevent data leakage — a common pitfall studies that can significantly compromise the validity of the results.

\section{Results\label{results}}

In this section, we present the outcomes of our machine learning models applied to satellite and ground-based data for estimating ground level CO\textsubscript{2} concentrations. We first compare the performance of different models, including Category Boosting, Extreme Gradient Boosting, and neural networks, in terms of their accuracy and robustness. This is followed by an analysis of multi-scale CO\textsubscript{2} predictions, extending beyond the ICOS station data to larger geographic areas. Finally, we evaluate the significance of various input features using SHAP analysis, providing insights into the key drivers of the model's predictions.

\subsection{Ground Level CO\texorpdfstring{\textsubscript{2}}{2} Estimation}
To perform the ground level CO\textsubscript{2} estimation, we used the baseline model, Category Boosting, Extreme Gradient Boosting, and neural network regression as described in Section \ref{sec:Models}. 

% Comparison and presentation of metrics
The baseline model achieved an Root Mean Squared Error (RMSE) of $6.22$. This is a measure of the amount of error in the predictions and a low RMSE is desirable. The Category Boosting model exhibited an RMSE of $5.14$, while the Extreme Gradient Boosting model, achieved an RMSE of $4.29$. Finally, the neural network achieved an RMSE of only $3.92$. Table \ref{tab:modelcomparison1} present a comprehensive comparison of these models, including the additional statistical metrics of Mean Squared Error (MSE) and Adjusted R2, providing a holistic view of model performances. These results suggest that the neural network is able to capture more of the variability in the data. These metrics are calculated using
\begin{equation}
\begin{array}{c}
\text{MSE} = \frac{1}{n}\sum_{i=1}^{n}(y_i - \hat{y}_i)^2, \hspace{0.3cm} \text{RMSE} = \sqrt{\mbox{MSE}}, \\
R^2_{\text{adj}}=1-\frac{n-1}{n-p-1}\cdot\frac{\displaystyle\sum_{i=1}^{n}(y_i - \hat{y}_i)^2}{\displaystyle\sum_{i=1}^{n}(y_i - \bar{y})^2},
\end{array}
\label{eq:metrics}
\end{equation}
where $n$ is the number of data points, $p$ is the number of input features, $y_i$ is the true ICOS-measurement, $\hat{y}_i$ is the estimation, and $\bar{y}$ is the average ICOS-measurement.

\begin{figure}[ht]
    \centering
    \includegraphics[width=0.6\linewidth,height=.25\linewidth]{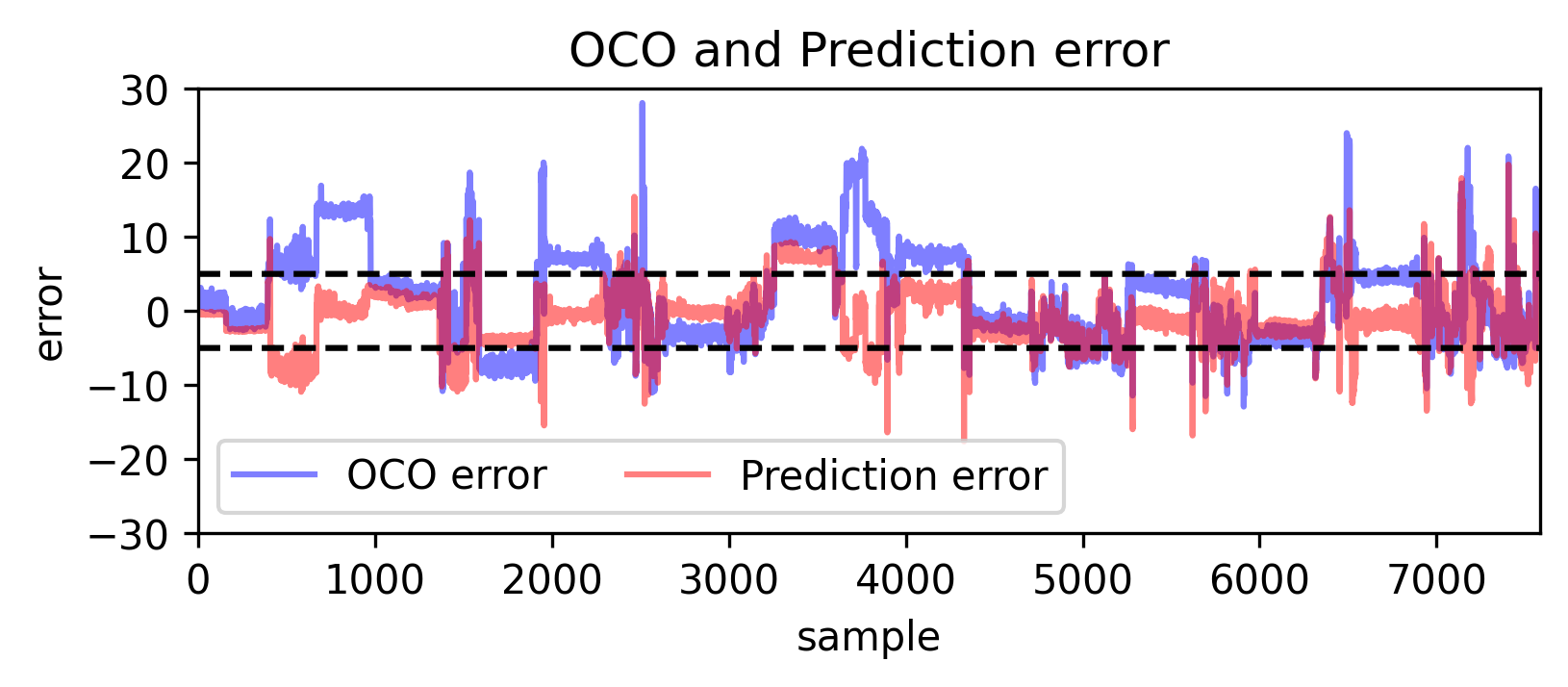}
    \caption{In blue, difference between the actual ICOS and actual OCO measurements. In red, difference between actual ICOS and predicted ICOS. Predictions are produced by the neural network regression model. The dashed lines highlight that most of the predicted values lie within an error of $\pm$5ppm. 
    }
\label{datavspred}
\end{figure}

\begin{table}[ht]
    \caption{Model comparisons based on metrics in (\ref{eq:metrics}). 
    }
    \centering
    \begin{tabular}{|l||l|l|l|l|l|}
        \hline
        \textbf{Model}             & \textbf{RMSE} & \textbf{MSE} & \textbf{adj.R2}  \\ 
        \hline
        \textit{Baseline}          
        & 6.22          & 38.7       & 0.516 \\ 
        \hline
        \textit{Category Boosting}
        & 5.14          & 26.4        & 0.671  \\ 
        \hline
        \textit{Gradient Boosting} 
        & 4.29        & 18.4        & 0.769 \\ 
        \hline
        \textit{Neural Regression}   
        & 3.92        & 15.3        & 0.808 \\ 
        \hline
    \end{tabular}
    \label{tab:modelcomparison1}
\end{table}
% Prediction errors + Effect of weather & location data
To tangibly compare the performance of the neural regression network to the baseline, Figure \ref{datavspred} displays the prediction errors from the two respective models. From the results it is clear that some patterns seen in the baseline are also visible in the neural network predictions. Mainly, Figure \ref{datavspred} shows clear flat sections where prediction errors remain semi-constant in both models. These are satellite measurements that are close in space and time and it is therefore natural that they result in similar predictions. However, it is also clear that the model is effectively able to shift the error of these sections closer to 0 using the additional contextual weather and location data provided.

\begin{figure}[h!]
    \centering
    \includegraphics[width=0.65\linewidth, height=.65\linewidth]{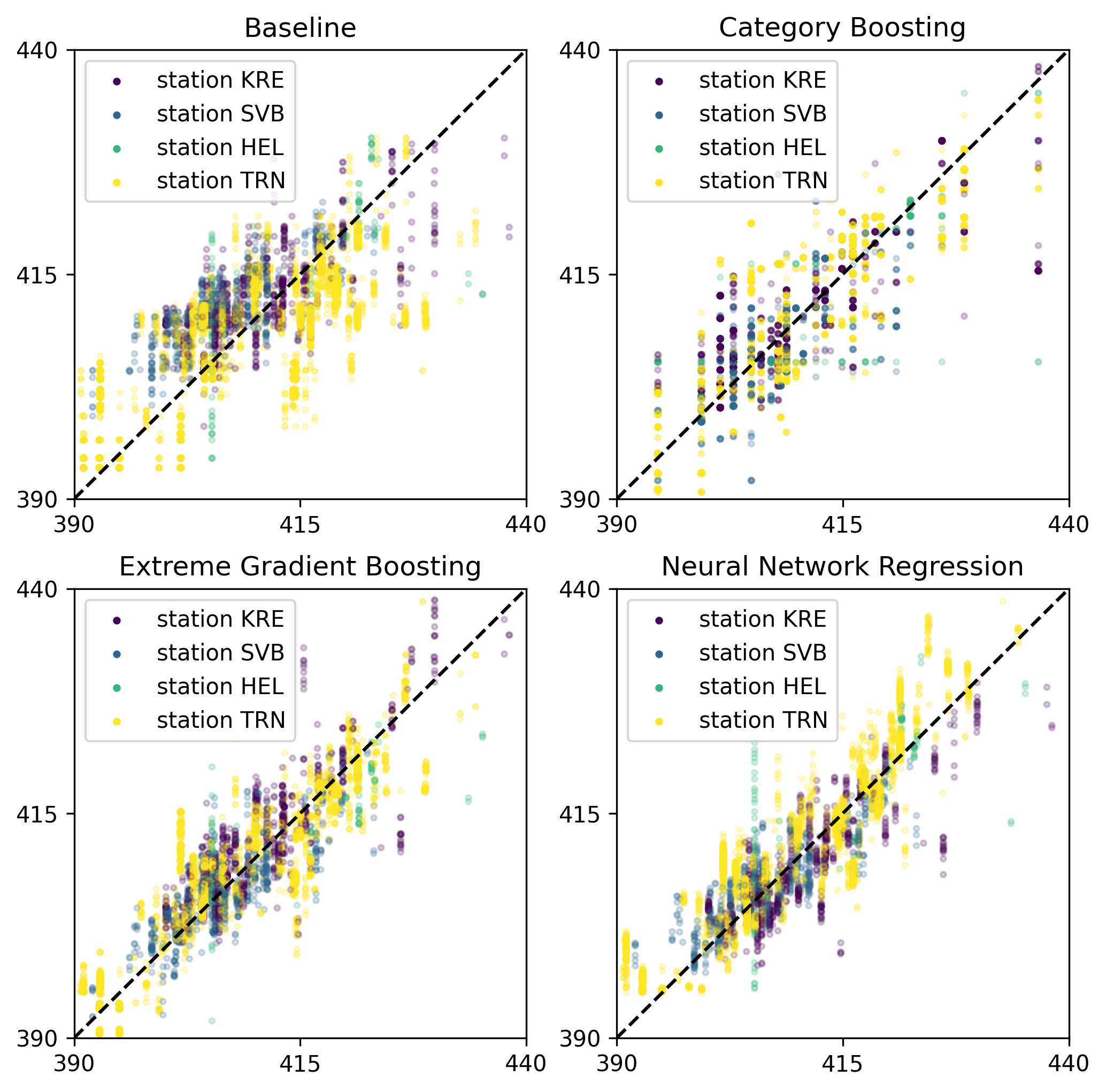}
    \caption{Correspondence plot between the predicted and true CO\textsubscript{2} levels for the test dataset. The x-axis shows the predicted CO\textsubscript{2} level from each respective model, while the y-axis shows to the true ICOS measurement. The locations for each station's acronym listed in the legend can be found at \url{https://www.icos-cp.eu/observations/station-network}. 
    }
    \label{prediction_plot}
\end{figure}

% Prediction vs ground truth + Category Boosting consequences
In addition, Figure \ref{prediction_plot} illustrates the predictions on the test set.  We can infer from this figure that the neural network predictions are somewhat better aligned with the true ICOS measurements, as indicated by their proximity to the diagonal line as well as the lower RMSE score of Table \ref{tab:modelcomparison1}. Moreover, the effects of the discretization in the Category Boosting model are apparent in the figure, where the predictions are constrained to a set of fixed values. This is a consequence of the architecture of the Category Boosting methods, which, instead of predicting a value for the ground level CO\textsubscript{2}, classifies the measurement as belonging to one of fixed set of possible values. This lowers the resolution of the predictions, but can sometimes be more effective at predictions. This was evidently not the true in this case.

\subsection{Multi-scale Predictions} \label{sec:multi-scale}

The results above are based on ground truth measurements from ICOS stations, which have a limited geographical reach. However, since the predictive features rely solely on satellite information from ERA5 and OCO satellites, predictions do not need to be confined to just ICOS locations. To do this for a given region and timescale, OCO measurements were collected and combined with the corresponding weather information. These data points were used to make estimates of ground level CO\textsubscript{2} at each point. Since OCO doesn't have full coverage every year, the available measurements have to be interpolated to neighboring locations.

We decided to interpolate the CO\textsubscript{2}\ levels via a weighted \textit{K-nearest neighbor} (KNN) interpolation method, where the weight function was chosen to be $w=1/d^p$, where $d$ is the distance to the measurement and $p$ is parameter corresponding to the decay in the weight function. This methodology estimates the value of CO\textsubscript{2}\ at a given location by considering the values at the $K$ nearest measured locations and assigning weights to them based on their distance from the point of interest.

Specifically, given a set of measurements $\{(x_i, y_i)\}_{i=1}^N$, where $x_i$ represents the location and $y_i$ is the measured CO\textsubscript{2}\ level at that location, the goal is to estimate the value $y^*$ at a new location $x^*$. The pseudocode of the proposed methodology is provided in Algorithm \ref{alg:interpolation}.

\begin{algorithm}[H]
\caption{Interpolation Procedure}\label{alg:interpolation}
\begin{algorithmic}[1]
%\STATE Identify the $K$ nearest neighbors $\{(x_{i_k}, y_{i_k})\}_{k=1}^K$ to the point $x^*$ based on Euclidean distance, $d(x^*, x_{i_k}) = \sqrt{\sum_{j=1}^d \left(x^*_j -x_{i_kj}\right)^2},$
\STATE Identify the $K$ nearest neighbors $\{(x_{i_k}, y_{i_k})\}_{k=1}^K$ to the point $x^*$ based on distance $d(x^*, x_{i_k})$ along the surface of the earth between the point $x^*$ and the $k$-th neighbor point $x_{i_k}$.

\STATE 
Compute the weight $w_k$ for each neighbor based on the distance from $x^*$, using the weight function
    \begin{equation}
    w_k = \frac{1}{d(x^*, x_{i_k})^p},
    \label{eq:weight}
    \end{equation}
    where $p$ is a parameter controlling the decay of the weight with distance.
\STATE
Normalize the weights to ensure they sum to 1:
    \(
    \tilde{w}_k = \frac{w_k}{\sum_{j=1}^K w_j},
    \)
    where $\tilde{w}_k$ is the normalized weight for the $k$-th neighbor.
\STATE
The interpolated value $y^*$ at the location $x^*$ is then computed as a weighted sum of the values at the nearest neighbors:
    \begin{equation}
    y^* = \sum_{k=1}^K \tilde{w}_k y_{i_k}, 
    \label{eq:ystar}
    \end{equation}
    where $y_{i_k}$ is the measured CO\textsubscript{2}\ level at the $k$-th nearest neighbor.
\end{algorithmic}
\end{algorithm}

\begin{figure}%[H]%[ht]
    \centering
    \includegraphics[width=\linewidth,height=.95\linewidth]{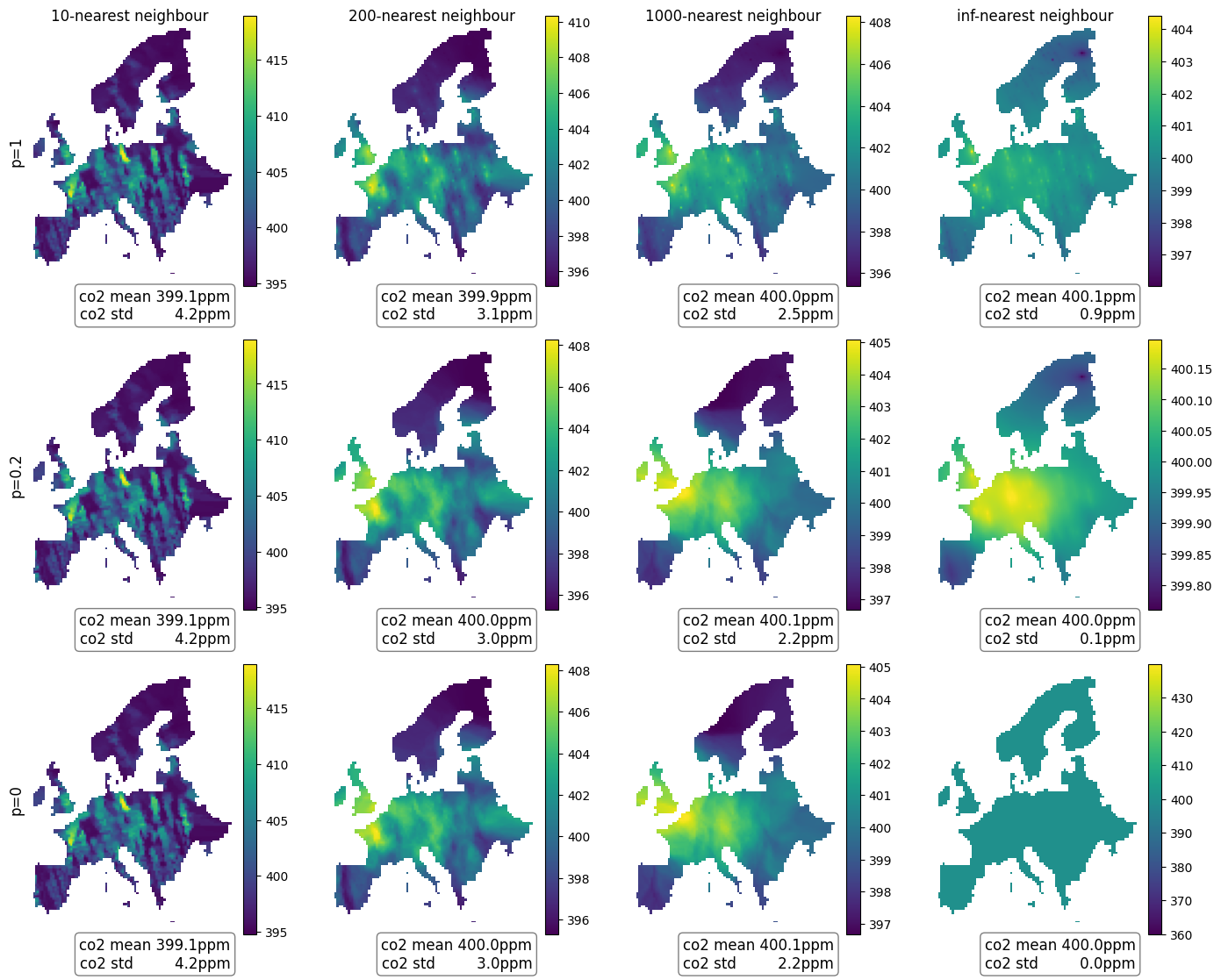}
    \caption{Interpolated predictions for Europe’s 2015 ground level CO\textsubscript{2}\ concentrations using different parameterizations of the weighted K-nearest neighbor interpolation. The ablation study shows results as the nearest neighbor parameter $K$ ranges from 10, 200, 1000, to $\infty$ (left to right), while the decay rate $p$ is varied through 1, 0.2, and 0 (top to bottom). See Algorithm \ref{alg:interpolation} for details.}
    \label{interpolation_effect}
\end{figure}

This methodology allows for a more flexible interpolation that adapts to the density and distribution of the measured data, with the parameters $K$ and $p$ providing control over the smoothness and locality of the interpolation.
The effect of these parameters can best be understood in the ablation study presented in Figure \ref{interpolation_effect}. It was observed that the average CO\textsubscript{2}\ levels remained relatively constant and close to the mean of the measurements, regardless of the interpolation parameters. However, the standard deviation varied significantly between parameterizations, decreasing as the values of both $K$ and $p$ in (\ref{eq:ystar}) and (\ref{eq:weight}) respectively, increased.

The parameter $K$ functions as a smoothing parameter, with a too low $K$ resulting in noise and artifacts from the satellite's orbit, and a too high $K$ removing a lot of the details in the predictions. The parameter $p$ has a small effect when $K$ is also low since measurements are fairly close and thus have a similar distance. However, as $K$ increases, the parameter $p$ has a large effect on the shape of the predictions. We found that using $K=200$ and $p=0.05$ resulted in visually reasonable results on the European scale, while $K=100$ and $p=0.05$ resulted in more salient values for small scale municipal estimations. As stated before, the lack of ground truth data means that these parameter setting are unverified. They are only used for the example regional visualizations expressed in this section.

\begin{figure}[h!]
    \centering
    \includegraphics[width=0.8\linewidth]{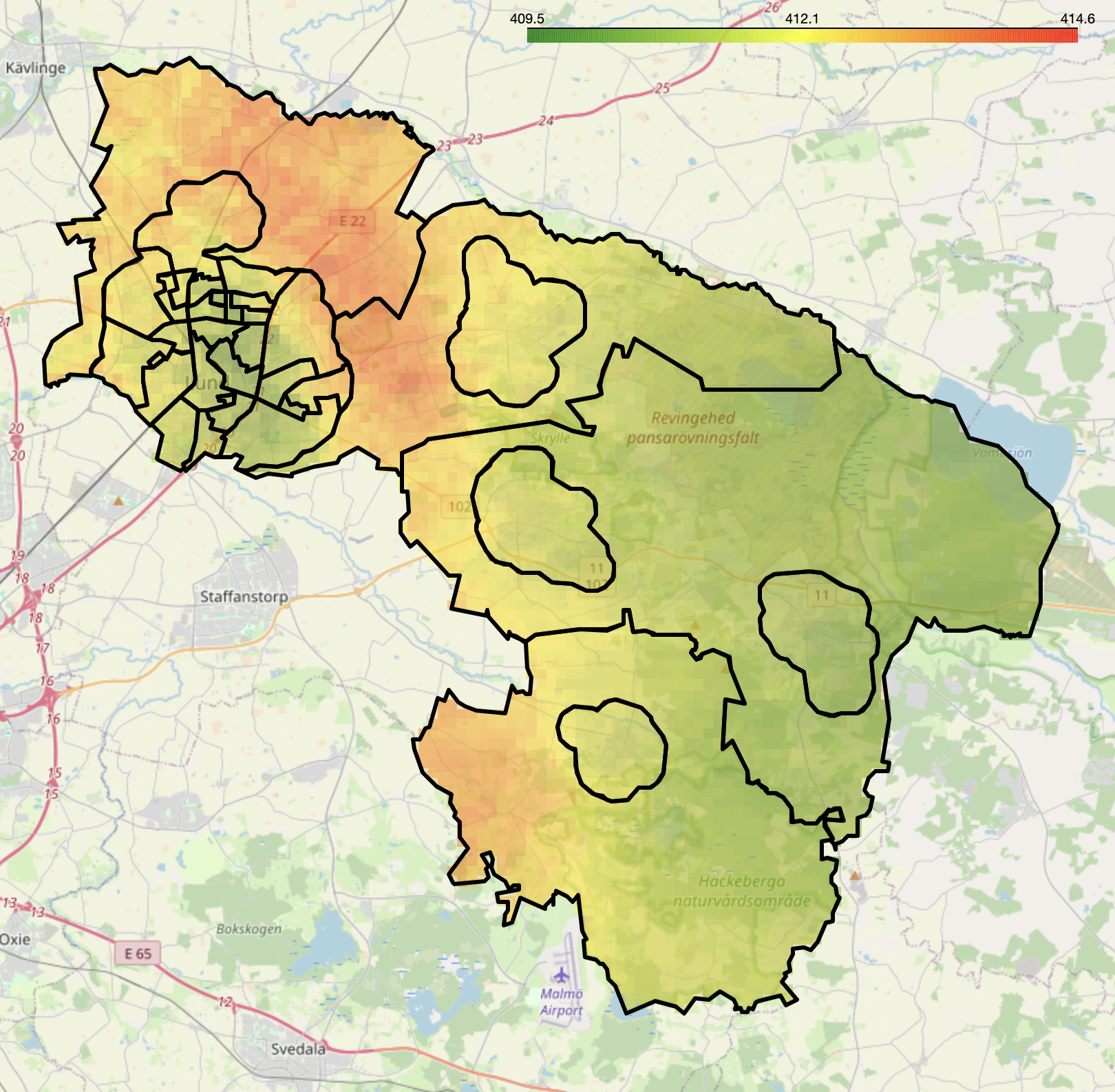}
    \caption{Predictions on the municipality of Lund, in southern Sweden, using the neural network regression model presented in Section \ref{sec:Models}. The KNN interpolation is made using $K=100$ and $p=0.05$. The estimated CO\textsubscript{2} levels are highest close to the city of Lund (top left) and further south near the airport (center bottom)  at the city of Malmö while they are lowest around the Häckeberga natural park reserve (middle right). }
    \label{municipality_heatmap}
\end{figure}

As a first example, Figure \ref{municipality_heatmap} displays the average predicted ground level CO\textsubscript{2} levels over Lund Municipality. The highest level of ground level CO\textsubscript{2} are predicted to be around northern Lund, and the lowest around Häckeberga nature preserve. We do not have ground truth data here to compare to, but at a glance this is a reasonable result as the preserve is reasonably a carbon sink. The higher levels to the north of Lund could possibly be due to the highway or perhaps the Örtofta sugar mill. Although it is important to stress that these results cannot be verified. 

\begin{figure}[tb]
    \centering
    \includegraphics[width=\linewidth]{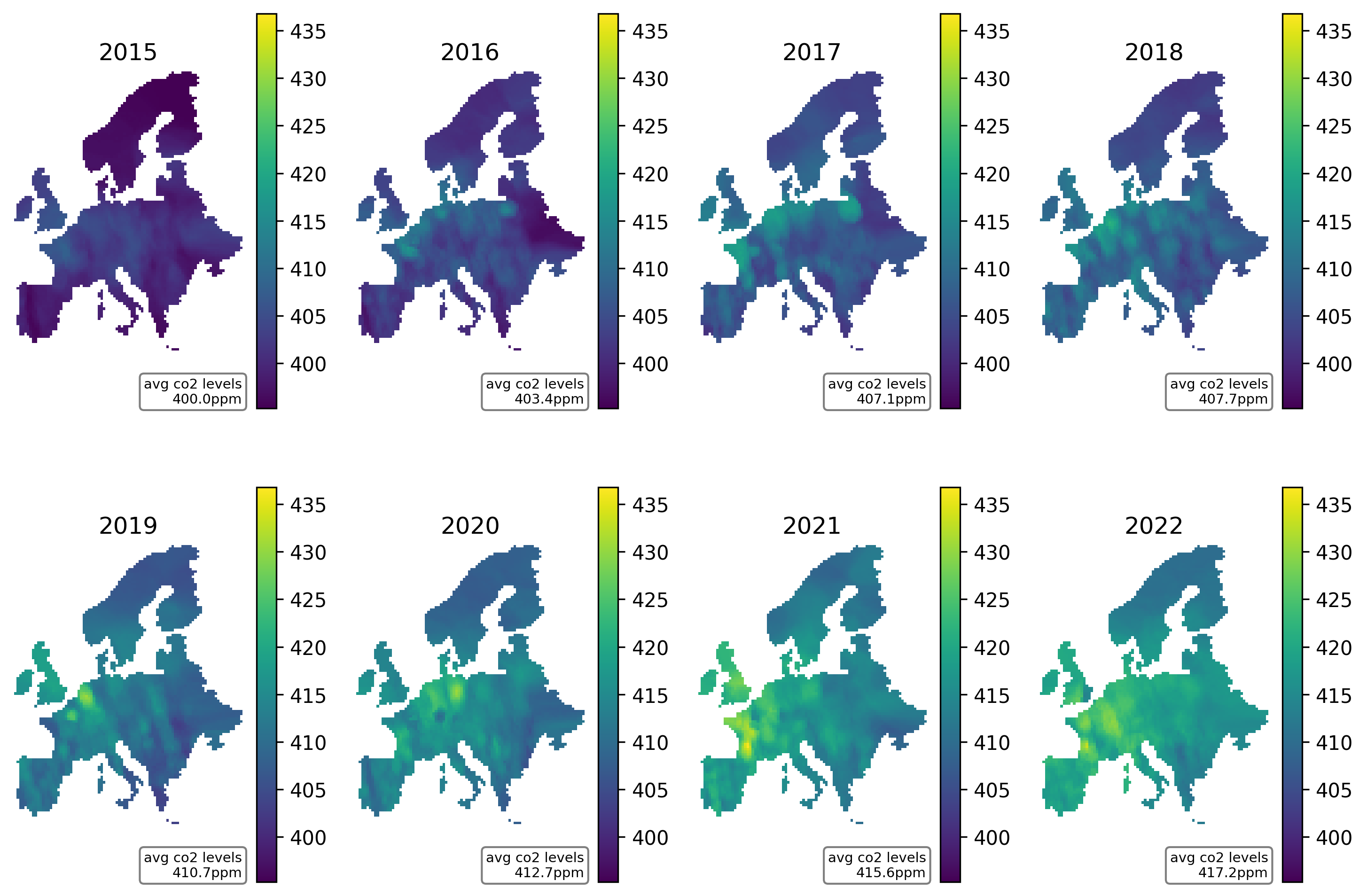}
    \caption{Annual ground level CO\textsubscript{2}\ estimations across Europe from 2015 to 2022, derived using the neural network regression model described in Section \ref{sec:Models}. The figure highlights spatial variations and temporal trends in CO\textsubscript{2}\ concentrations over the entire region, showcasing the model's capability to capture and predict CO\textsubscript{2}\ levels at a high spatial resolution. The KNN interpolation was made using $K=200$ and $p=0.05$.}
    \label{fig:europe_heatmap}
  \end{figure}

The proposed approach has significant potential for scaling to larger regions, enabling the generation of high-resolution predictions across both time and space. Figure \ref{fig:europe_heatmap} illustrates yearly ground level CO\textsubscript{2}\ estimations across Europe, capturing the expected upward trend in CO\textsubscript{2}\ concentrations over time. Notably, these predictions are produced at a high spatial resolution, even for locations where no ground stations currently measure CO\textsubscript{2}. Although the limited availability of ground level CO\textsubscript{2}\ data at such high spatial resolutions restricts our ability to independently validate these results, the figures highlight the model's versatility. The model is capable of providing detailed predictions across varying spatial and temporal scales, surpassing the resolution of existing ground level measurements.

\subsection{Feature Importance}
% Feature Importance
To analyze the importance of the OCO and weather features used in predictions, a SHAP analysis \cite{lundberg2017shap} was employed. This method measures the change in output from the model when an input feature is changed. The feature importance is the amplitude of the prediction change when tweaking the corresponding feature. 

\begin{figure}[h!]
    \centering
    \includegraphics[width=.8\linewidth]{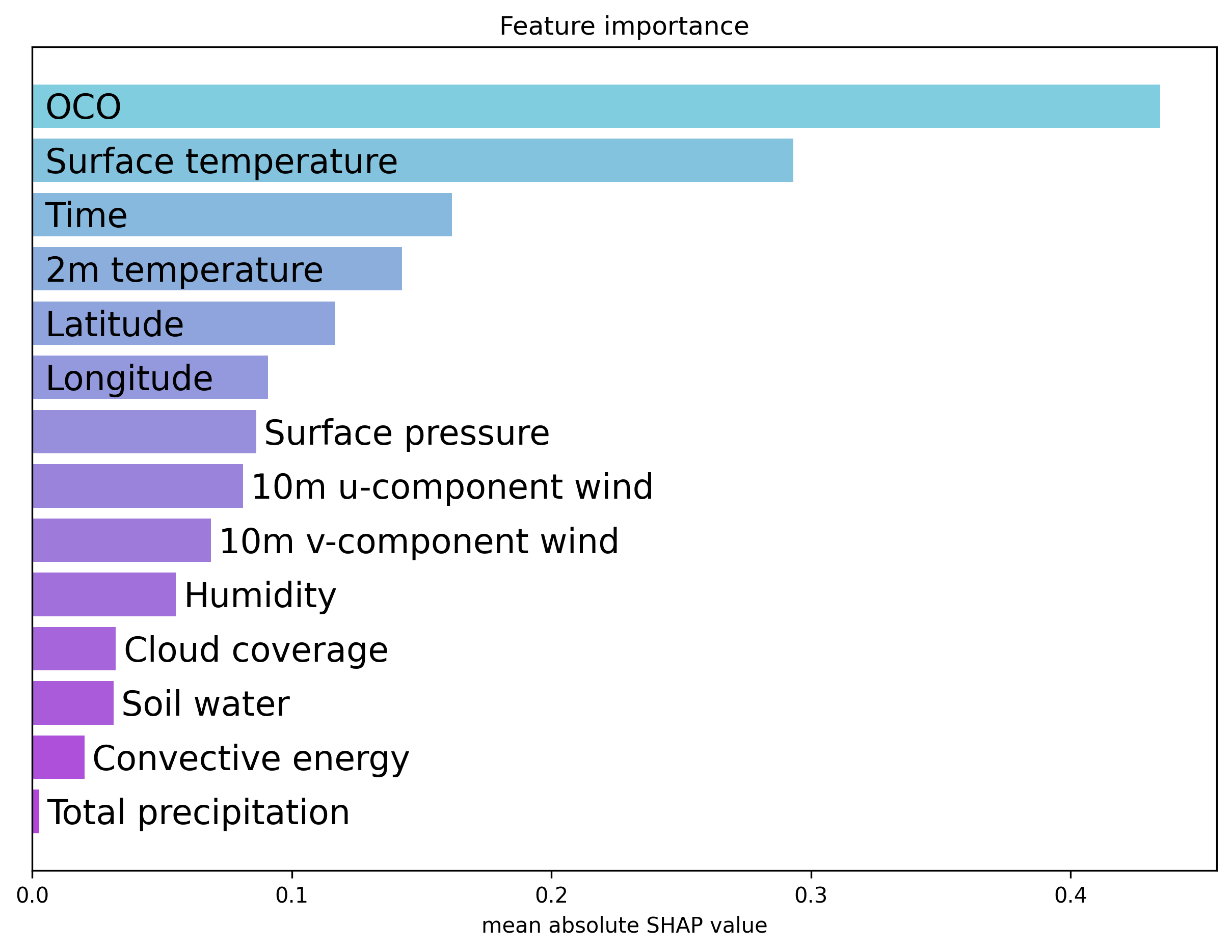}
\caption{The importance of features in the input data using SHAP analysis based on the model results presented in Figure \ref{datavspred}. The three most important features were the xCO\textsubscript{2} OCO measurements, surface temperature and time. The colors in the figure are used solely for differentiation and do not convey any additional information. }
\label{featureImp}
\end{figure}

The results, shown in Figure \ref{featureImp}, highlight the relative importance of various features in our predictive model. The most significant feature identified was the xCO\textsubscript{2} from OCO2. This finding is expected, as xCO\textsubscript{2} provides a direct measurement of carbon concentration, making it a crucial indicator for our analysis.

In addition to xCO\textsubscript{2}, temperature variables, specifically surface temperature and temperature at 2 meters, were also identified as important features. This result could be attributed to the fact that temperature serves as a proxy for the vertical movement of CO\textsubscript{2}, influencing its distribution and concentration in the atmosphere, see \cite{LSW2017, GeRo} for an overview.
Moreover, the analysis revealed that geographic location (latitude and longitude) and temporal variables (time) are significant predictors. These features likely capture the global variations in atmospheric patterns that are not fully explained by other weather-related variables. Including location and time in the model helps account for the spatial and temporal heterogeneity in atmospheric CO\textsubscript{2} levels.

Overall, the SHAP analysis underscores the necessity of incorporating a diverse set of features, including direct measurements, proxy indicators, and spatial-temporal variables, to enhance the predictive accuracy of atmospheric CO\textsubscript{2} models.

\section{Discussion}

The results indicate that we can achieve accurate predictions of ground level CO\textsubscript{2} with an RMSE of under 4ppm and an adjusted R2 value of over 80\%. This can roughly be interpreted as the model typically guessing within 4ppm of the true value and that the model can explain over 80\% of the variation in ground level CO\textsubscript{2}. 

Since the predictive features only consists of satellite data, the model can be applied to locations outside of ICOS stations and is therefore able to render high resolution predictions for any region. However, it is important to note that since ground truth data does not exist outside of ICOS stations, the extrapolated results cannot be verified at this moment. 

% Limitation 1: ICOS bias
The primary limitation of the current iteration of the model is the lack of good ground truth data. Since ground truth data is sparse and only localized around ICOS stations, the model will likely contain some biases towards these locations. Due to their size, stations are predominantly placed in rural areas or uninhabited regions, this means that there is very little ground truth data in cities or highly industrialized regions. In addition to this rural bias, we also have a continental bias as the ICOS stations are only located in Europe. The vast variety of landscapes, weather, and emission patterns that exist in the world cannot be said to be captured in the localized ground truth data available. Therefore, while the global reach of satellites allow for world-wide use of the model, it is highly discouraged at this stage. The primary application, for now, remains inside Europe. 

That being said, the model can easily be adapted with additional ground truth data from any location in the world, thus opening up the possibility for global use in future iterations. 

% Limitation 2: OCO bias
The second limitation is that while OCO has global reach, it doesn't produce evenly distributed measurements either in space or time. Some regions, such as the south of Europe, receive much more observations compared to northern Europe due to their geographical location. Closer to the poles, the observations are sparser and only receive a few orbits per year. This means that weekly or monthly averages will be highly inaccurate for many locations. Therefore, analyses should ideally keep to yearly averages or larger time spans.

% Primary use of the model with limitations in mind and where we future reseach should improve
With this in mind, the current stage of the model is primarily suited for estimations in Europe of yearly averages in large scale regions, or multi-year averages in smaller areas. The high level resolution of the prediction could give additional insights into where emissions stem from and where the most important carbon sinks are located. To increase usability, future research should focus mainly on expanding the availability of good ground truth data. While the ICOS stations provide invaluable information with great resolution in time, they provide only part of the picture due to their limited reach. A dataset consisting of ground level CO\textsubscript{2} measurements with low time resolution but high spatial spread would be a great contribution to this field as a complement to ICOS. Incorporation of global ground level CO\textsubscript{2} measurements would also contribute greatly by ensuring global reach and availability of predictions.

This methodology of using remote sensing with verified ground truth data will be crucial for equitable and informed climate action due to the fact that manual emission testing or CO\textsubscript{2} measuring stations are not an option for everyone. If this is done with steps ensuring that the predictions are unbiased and representative of most locations and conditions, we believe that this can be a vital tool for local effort to mitigate climate change.

\subsection*{Towards better climate action}
%\subsection*{Enhancing Localized CO\textsubscript{2} Monitoring for Targeted Climate Action}

The foundation of our research lies in addressing the significant gap in localized CO\textsubscript{2} measurements, which hampers the effective design, implementation, and monitoring of precise local climate initiatives. The remote sensing approach we present has demonstrated its capacity to predict local CO\textsubscript{2} levels, offering a promising alternative to this critical shortcoming. At the same time, we outline a clear pathway for further refinement of this approach, emphasizing the need to ensure that verified ground truth data is unbiased and representative of diverse global locations and conditions.

Given the currently available data, the model introduced in this article holds significant potential to enhance climate action efforts by enabling:

\begin{itemize} 
\item Identification of High CO\textsubscript{2} Concentration Areas: This allows for the strategic targeting of climate actions where they are most needed, optimizing resource allocation. 
\item Monitoring of Temporal Changes in CO\textsubscript{2} Levels: This facilitates the assessment of climate action effectiveness over time, providing feedback that can guide future interventions. 
\item Effective Public Communication: By using visual tools such as maps and graphs, we can communicate CO\textsubscript{2} level changes clearly to the public, fostering greater engagement and commitment to climate initiatives. 
\item Validation of Reported Emission Data: This contributes to the enforcement of environmental policies by verifying the accuracy of reported emissions, enhancing transparency and accountability. 
\item Support for Multi-Level Governance: Local administrations and European policymakers can rely on the same data sources, promoting coordinated and cohesive climate strategies across different governance levels. 
\item Integration with Socio-Economic Data: Combining local CO\textsubscript{2} data with socio-economic indicators enables the analysis of synergies and trade-offs within socio-ecological systems, supporting more holistic decision-making. 
\end{itemize}
Due to the limitations of verified ground truth available, the current model is particularly well-suited for use across Europe, providing valuable insights for large-scale regions (such as countries) based on yearly averages, as well as for smaller-scale regions (such as municipalities) when considering multi-year averages.

\section{Conclusions\label{conclusions}}
Atmospheric CO\textsubscript{2} monitoring has seen substantial improvements in recent years, driven by advancements in technology. However, significant challenges persist. The commonly used bottom-up approaches, which rely heavily on reported data, often suffer from biases and gaps in data coverage. Moreover, the long atmospheric lifespan of CO\textsubscript{2} compared to other pollutants like methane and aerosols leads to extensive spatial distribution due to atmospheric transport. To address these challenges, the integration of detailed weather data with sophisticated atmospheric models is crucial. As satellite technology continues to evolve, it offers increasing potential for more precise and localized monitoring of CO\textsubscript{2} emissions, which is critical for effective climate action and informed policy-making.

Our study aims to provide a more granular view of CO\textsubscript{2} emissions to support policymakers in developing precise local climate action strategies. The proposed methodology goes beyond traditional approaches that rely on national data downsampling, by leveraging multimodal data collection for more accurate results.

%Limitations
While the proposed methodology shows significant promise, it does have some limitations. A potential limitation of the proposed methodology is the sparse availability of ground truth CO\textsubscript{2} measurements, particularly in urban and industrial regions. The reliance on data from ICOS stations, which are primarily located in rural and European settings, may introduce biases, potentially underrepresenting CO\textsubscript{2} variability in areas with higher emissions. Additionally, the OCO-2 and OCO-3 satellites do not provide uniform temporal and spatial coverage across all regions, particularly in areas with complex terrain or those distant from the equator. This uneven coverage could result in data gaps and impact the accuracy of CO\textsubscript{2} estimations over time and space, particularly in regions with limited direct observations.

%Future Directions
The integration of satellite data with ground measurements is shown to be crucial for understanding and mitigating CO\textsubscript{2} emissions at a local level. This approach not only enhances precision in emissions monitoring but also lays a solid foundation for policy development, community engagement, and collaborative efforts towards global climate goals. Continued advancements in satellite technology, such as the upcoming Copernicus CO\textsubscript{2} Monitoring Mission (CO2M), promise to enhance the spatial and temporal resolution of CO\textsubscript{2} data, enabling more accurate and frequent monitoring of localized emissions \cite{future}. Future research should prioritize improving the accuracy of satellite-based CO\textsubscript{2} measurements, expanding the availability of ground level CO\textsubscript{2} data and integrating comprehensive meteorological datasets. These advancements will be key to deepening our understanding and management of CO\textsubscript{2} emissions on both regional and global scales.

\section*{Acknowledgments}
This work is partially supported by FORMAS no. 2022-151862. The work of O.~Å. and A.~S. is also partially supported by grants from eSSENCE no. 138227 and Rymdstyrelsen no. 2022-00282.  The computations were enabled by resources provided by the National Academic Infrastructure for Supercomputing in Sweden (NAISS), partially funded by the Swedish Research Council through grant agreement no. 2022-06725.

\vskip 0.2in
\bibliography{biblioICOS}

\begin{thebibliography}{28}
\providecommand{\natexlab}[1]{#1}
\providecommand{\url}[1]{\texttt{#1}}
\expandafter\ifx\csname urlstyle\endcsname\relax
  \providecommand{\doi}[1]{doi: #1}\else
  \providecommand{\doi}{doi: \begingroup \urlstyle{rm}\Url}\fi

\bibitem[OCO(2023)]{OCO2DataCenter}
Oco-2 data center, 2023.
\newblock URL \url{https://ocov2.jpl.nasa.gov}.
\newblock Accessed: 2023-12-30.

\bibitem[Beirle et~al.(2011)Beirle, Boersma, Platt, Lawrence, and Wagner]{Beirle2011}
S.~Beirle, K.~Folkert Boersma, Ulrich Platt, Mark~G. Lawrence, and Thomas Wagner.
\newblock Megacity emissions and lifetimes of nitrogen oxides probed from space.
\newblock \emph{Science}, 333\penalty0 (6050):\penalty0 1737--1739, 2011.
\newblock \doi{10.1126/science.1207824}.
\newblock URL \url{https://www.science.org/doi/abs/10.1126/science.1207824}.

\bibitem[Bovensmann et~al.(2010)Bovensmann, Buchwitz, Burrows, Reuter, Krings, Gerilowski, Schneising, Heymann, Tretner, and Erzinger]{Bovensmann2010}
H.~Bovensmann, M.~Buchwitz, J.~P. Burrows, M.~Reuter, T.~Krings, K.~Gerilowski, O.~Schneising, J.~Heymann, A.~Tretner, and J.~Erzinger.
\newblock A remote sensing technique for global monitoring of power plant co$_{2}$ emissions from space and related applications.
\newblock \emph{Atmospheric Measurement Techniques}, 3\penalty0 (4):\penalty0 781--811, 2010.
\newblock \doi{10.5194/amt-3-781-2010}.
\newblock URL \url{https://amt.copernicus.org/articles/3/781/2010/}.

\bibitem[Cai et~al.(2020)Cai, Xu, Zhu, Hu, and Li]{cai2020prediction}
Jianchao Cai, Kai Xu, Yanhui Zhu, Fang Hu, and Liuhuan Li.
\newblock Prediction and analysis of net ecosystem carbon exchange based on gradient boosting regression and random forest.
\newblock \emph{Applied energy}, 262:\penalty0 114566, 2020.

\bibitem[Claeyman et~al.(2011)Claeyman, Atti\'e, Peuch, El~Amraoui, Lahoz, Josse, Ricaud, von Clarmann, H\"opfner, Orphal, Flaud, Edwards, Chance, Liu, Pasternak, and Canti\'e]{Claeyman2011}
M.~Claeyman, J.-L. Atti\'e, V.-H. Peuch, L.~El~Amraoui, W.~A. Lahoz, B.~Josse, P.~Ricaud, T.~von Clarmann, M.~H\"opfner, J.~Orphal, J.-M. Flaud, D.~P. Edwards, K.~Chance, X.~Liu, F.~Pasternak, and R.~Canti\'e.
\newblock A geostationary thermal infrared sensor to monitor the lowermost troposphere: O$_{3}$ and co retrieval studies.
\newblock \emph{Atmospheric Measurement Techniques}, 4\penalty0 (2):\penalty0 297--317, 2011.
\newblock \doi{10.5194/amt-4-297-2011}.
\newblock URL \url{https://amt.copernicus.org/articles/4/297/2011/}.

\bibitem[Eldering et~al.(2017)Eldering, Wennberg, Crisp, Schimel, Gunson, Chatterjee, Liu, Schwandner, Sun, O’dell, et~al.]{Eldering2017}
A~Eldering, PO~Wennberg, D~Crisp, DS~Schimel, MR~Gunson, A~Chatterjee, J~Liu, FM~Schwandner, Y~Sun, CW~O’dell, et~al.
\newblock The orbiting carbon observatory-2 early science investigations of regional carbon dioxide fluxes.
\newblock \emph{Science}, 358\penalty0 (6360):\penalty0 eaam5745, 2017.

\bibitem[Feldman et~al.(2023)Feldman, Zhang, Yoshida, Chatterjee, and Poulter]{Feldman2023}
A.~F. Feldman, Z.~Zhang, Y.~Yoshida, A.~Chatterjee, and B.~Poulter.
\newblock Using orbiting carbon observatory-2 (oco-2) column co2 retrievals to rapidly detect and estimate biospheric surface carbon flux anomalies.
\newblock \emph{Atmospheric Chemistry and Physics}, 23\penalty0 (2):\penalty0 1545--1563, 2023.
\newblock \doi{10.5194/acp-23-1545-2023}.
\newblock URL \url{https://acp.copernicus.org/articles/23/1545/2023/}.

\bibitem[Fioletov et~al.(2015)Fioletov, McLinden, Krotkov, and Li]{fioletov2015lifetimes}
V.~E. Fioletov, C.~A. McLinden, N.~Krotkov, and C.~Li.
\newblock Lifetimes and emissions of so2 from point sources estimated from omi.
\newblock \emph{Geophysical Research Letters}, 42\penalty0 (6):\penalty0 1969--1976, 2015.

\bibitem[Geldhauser and Romito(2019)]{GeRo}
Carina Geldhauser and Marco Romito.
\newblock The point vortex model for the {E}uler equation.
\newblock \emph{AIMS Math.}, 4\penalty0 (3):\penalty0 534--575, 2019.
\newblock ISSN 2473-6988.
\newblock \doi{10.3934/math.2019.3.534}.
\newblock URL \url{https://doi.org/10.3934/math.2019.3.534}.

\bibitem[Hakkarainen et~al.(2023)Hakkarainen, Ialongo, Szelag, and Oda]{Hakkarainen2023}
J.~Hakkarainen, I.~Ialongo, M.~E. Szelag, and T.~Oda.
\newblock Characterizing major anthropogenic point sources in the south african highveld region using oco-3 carbon dioxide snapshot area maps and sentinel-5p/tropomi nitrogen dioxide columns.
\newblock \emph{IOP Science}, 2023.
\newblock URL \url{https://iopscience.iop.org/article/10.1088/1748-9326/ac9f58}.

\bibitem[He et~al.(2018)He, Lei, Welp, Zeng, Bie, Yang, and Liu]{he2018satellite}
Z.~He, L.~Lei, L.~R Welp, Z.-C. Zeng, N.~Bie, S.~Yang, and L.~Liu.
\newblock Detection of spatiotemporal extreme changes in atmospheric co2 concentration based on satellite observations.
\newblock \emph{Remote Sensing}, 10\penalty0 (6):\penalty0 839, 2018.

\bibitem[Janardanan et~al.(2016)Janardanan, Maksyutov, Oda, Saito, Kaiser, Ganshin, Stohl, Matsunaga, Yoshida, and Yokota]{janardanan2016comparing}
Rajesh Janardanan, Shamil Maksyutov, Tomohiro Oda, Makoto Saito, Johannes~W Kaiser, Alexander Ganshin, Andreas Stohl, Tsuneo Matsunaga, Yukio Yoshida, and Tatsuya Yokota.
\newblock Comparing gosat observations of localized co2 enhancements by large emitters with inventory-based estimates.
\newblock \emph{Geophysical Research Letters}, 43\penalty0 (7):\penalty0 3486--3493, 2016.
\newblock \doi{10.1002/2016GL067843}.

\bibitem[{Keppel-Aleks} et~al.(2013){Keppel-Aleks}, {Wennberg}, {O'Dell}, and {Wunch}]{Keppel-Aleks2013}
G.~{Keppel-Aleks}, P.~O. {Wennberg}, C.~W. {O'Dell}, and D.~{Wunch}.
\newblock {Towards constraints on fossil fuel emissions from total column carbon dioxide}.
\newblock \emph{Atmospheric Chemistry \& Physics}, 13\penalty0 (8):\penalty0 4349--4357, April 2013.
\newblock \doi{10.5194/acp-13-4349-201310.5194/acpd-12-29887-2012}.

\bibitem[Kort et~al.(2012)Kort, Frankenberg, Costigan, Lindenmaier, Dubey, and Wunch]{kort2012four}
Eric~A Kort, Christian Frankenberg, Keeley~R Costigan, Rodica Lindenmaier, Manvendra~K Dubey, and Debra Wunch.
\newblock Four corners: The largest us methane anomaly viewed from space.
\newblock \emph{Geophysical Research Letters}, 39\penalty0 (18), 2012.
\newblock \doi{10.1029/2012GL052303}.

\bibitem[Liang et~al.(2017)Liang, Spall, and Wunsch]{LSW2017}
Xinfeng Liang, Michael Spall, and Carl Wunsch.
\newblock Global ocean vertical velocity from a dynamically consistent ocean state estimate.
\newblock \emph{Journal of Geophysical Research: Oceans}, 122\penalty0 (10):\penalty0 8208--8224, 2017.
\newblock \doi{https://doi.org/10.1002/2017JC012985}.
\newblock URL \url{https://agupubs.onlinelibrary.wiley.com/doi/abs/10.1002/2017JC012985}.

\bibitem[Lundberg and Lee(2017)]{lundberg2017shap}
Scott~M Lundberg and Su-In Lee.
\newblock A unified approach to interpreting model predictions.
\newblock \emph{Advances in neural information processing systems}, 30, 2017.

\bibitem[O'Dell et~al.(2018)O'Dell, Eldering, Wennberg, Crisp, Gunson, Fisher, Frankenberg, Kiel, Lindqvist, Mandrake, Merrelli, Natraj, Nelson, Osterman, Payne, Taylor, Wunch, Drouin, Oyafuso, Chang, McDuffie, Smyth, Baker, Basu, Chevallier, Crowell, Feng, Palmer, Dubey, Garc\'{\i}a, Griffith, Hase, Iraci, Kivi, Morino, Notholt, Ohyama, Petri, Roehl, Sha, Strong, Sussmann, Te, Uchino, and Velazco]{ODell2018}
C.~W. O'Dell, A.~Eldering, P.~O. Wennberg, D.~Crisp, M.~R. Gunson, B.~Fisher, C.~Frankenberg, M.~Kiel, H.~Lindqvist, L.~Mandrake, A.~Merrelli, V.~Natraj, R.~R. Nelson, G.~B. Osterman, V.~H. Payne, T.~E. Taylor, D.~Wunch, B.~J. Drouin, F.~Oyafuso, A.~Chang, J.~McDuffie, M.~Smyth, D.~F. Baker, S.~Basu, F.~Chevallier, S.~M.~R. Crowell, L.~Feng, P.~I. Palmer, M.~Dubey, O.~E. Garc\'{\i}a, D.~W.~T. Griffith, F.~Hase, L.~T. Iraci, R.~Kivi, I.~Morino, J.~Notholt, H.~Ohyama, C.~Petri, C.~M. Roehl, M.~K. Sha, K.~Strong, R.~Sussmann, Y.~Te, O.~Uchino, and V.~A. Velazco.
\newblock Improved retrievals of carbon dioxide from orbiting carbon observatory-2 with the version 8 acos algorithm.
\newblock \emph{Atmospheric Measurement Techniques}, 11\penalty0 (12):\penalty0 6539--6576, 2018.
\newblock \doi{10.5194/amt-11-6539-2018}.
\newblock URL \url{https://amt.copernicus.org/articles/11/6539/2018/}.

\bibitem[Peylin et~al.(2013)Peylin, Law, Gurney, Chevallier, Jacobson, Maki, Niwa, Patra, Peters, Rayner, R\"odenbeck, van~der Laan-Luijkx, and Zhang]{peylin2013}
P.~Peylin, R.~M. Law, K.~R. Gurney, F.~Chevallier, A.~R. Jacobson, T.~Maki, Y.~Niwa, P.~K. Patra, W.~Peters, P.~J. Rayner, C.~R\"odenbeck, I.~T. van~der Laan-Luijkx, and X.~Zhang.
\newblock Global atmospheric carbon budget: results from an ensemble of atmospheric co$_{2}$ inversions.
\newblock \emph{Biogeosciences}, 10\penalty0 (10):\penalty0 6699--6720, 2013.
\newblock \doi{10.5194/bg-10-6699-2013}.
\newblock URL \url{https://bg.copernicus.org/articles/10/6699/2013/}.

\bibitem[Reuter et~al.(2014)Reuter, Buchwitz, Hilboll, Richter, Schneising, Hilker, Heymann, Bovensmann, and Burrows]{reuter2014remote}
M~Reuter, M~Buchwitz, A~Hilboll, A~Richter, O~Schneising, M~Hilker, J~Heymann, H~Bovensmann, and JP~Burrows.
\newblock Decreasing emissions of nox relative to co2 in east asia inferred from satellite observations.
\newblock \emph{Nature Geoscience}, 7\penalty0 (11):\penalty0 792--795, 2014.
\newblock \doi{10.1038/ngeo2257}.

\bibitem[Reuter et~al.(2019)Reuter, Buchwitz, Schneising, Krautwurst, O'Dell, Richter, Bovensmann, and Burrows]{future}
M.~Reuter, M.~Buchwitz, O.~Schneising, S.~Krautwurst, C.~W. O'Dell, A.~Richter, H.~Bovensmann, and J.~P. Burrows.
\newblock Towards monitoring localized co\textsubscript(2) emissions from space: co-located regional co\textsubscript(2) and no\textsubscript(2) enhancements observed by the oco-2 and s5p satellites.
\newblock \emph{Atmospheric Chemistry and Physics}, 19\penalty0 (14):\penalty0 9371--9383, 2019.
\newblock \doi{10.5194/acp-19-9371-2019}.
\newblock URL \url{https://acp.copernicus.org/articles/19/9371/2019/}.

\bibitem[Schneising et~al.(2014)Schneising, Burrows, Dickerson, Buchwitz, Reuter, and Bovensmann]{schneising2013remote}
Oliver Schneising, John~P Burrows, Russell~R Dickerson, Michael Buchwitz, Maximilian Reuter, and Heinrich Bovensmann.
\newblock Remote sensing of fugitive methane emissions from oil and gas production in north american tight geologic formations.
\newblock \emph{Earth's Future}, 2\penalty0 (10), 2014.
\newblock \doi{10.1002/2014EF000265}.

\bibitem[Taylor et~al.(2023)]{Taylor2023}
T.~E. Taylor et~al.
\newblock Evaluating the consistency between oco-2 and oco-3 xco2 estimates derived from the nasa acos version 10 retrieval algorithm.
\newblock \emph{Atmospheric Measurement Techniques}, 16:\penalty0 3173--3209, 2023.
\newblock \doi{10.5194/amt-16-3173-2023}.

\bibitem[Team et~al.(2020)Team, Gunson, and Eldering]{OCO2_L2_Lite_FP_10r}
OCO-2~Science Team, Michael Gunson, and Annmarie Eldering.
\newblock \url{https://doi.org/10.5067/E4E140XDMPO2}, December 2020.
\newblock URL \url{https://disc.gsfc.nasa.gov/datasets/OCO2_L2_Lite_FP_10r/summary}.
\newblock Greenbelt, MD, USA, Goddard Earth Sciences Data and Information Services Center (GES DISC).

\bibitem[Wang et~al.(2013)Wang, Tao, Ciais, Shen, Huang, Chen, Shen, Wang, Zhang, Lu, Zhu, Chen, Liu, Wang, Wang, Liu, Li, and Piao]{wang2013}
R.~Wang, S.~Tao, P.~Ciais, H.~Z. Shen, Y.~Huang, H.~Chen, G.~F. Shen, W.~Wang, B.~Li, Y.~Y. Zhang, Y.~Lu, D.~Zhu, Y.~C. Chen, X.~P. Liu, W.~T. Wang, X.~L. Wang, W.~X. Liu, B.~G. Li, and S.~L. Piao.
\newblock High-resolution mapping of combustion processes and implications for co2 emissions.
\newblock \emph{Atmos. Chem. Phys.}, 13:\penalty0 5189--–5203, 2013.
\newblock \doi{10.5194/acp-13-5189-2013}.

\bibitem[Wang et~al.(2022)Wang, He, Feng, and Jin]{Wang2022}
W.~Wang, J.~He, H.~Feng, and Z.~Jin.
\newblock High-coverage reconstruction of xco2 using multisource satellite remote sensing data in beijing–tianjin–hebei region.
\newblock \emph{International Journal of Environmental Research and Public Health}, 19\penalty0 (17), 2022.
\newblock ISSN 1660-4601.
\newblock \doi{10.3390/ijerph191710853}.
\newblock URL \url{https://www.mdpi.com/1660-4601/19/17/10853}.

\bibitem[Wen et~al.(2021)Wen, Lu, and Jhang]{wen2021features}
Hung-Ta Wen, Jau-Huai Lu, and Deng-Siang Jhang.
\newblock Features importance analysis of diesel vehicles’ nox and co2 emission predictions in real road driving based on gradient boosting regression model.
\newblock \emph{International Journal of Environmental Research and Public Health}, 18\penalty0 (24):\penalty0 13044, 2021.

\bibitem[Ye et~al.(2020)]{Ye2020}
Xinxin Ye et~al.
\newblock Constraining fossil fuel co2 emissions from urban area using oco-2 observations of total column co2.
\newblock \emph{Journal of Geophysical Research: Atmospheres}, 125, 2020.
\newblock \doi{10.1029/2019JD030528}.

\bibitem[Yokota et~al.(2009)Yokota, Yoshida, Eguchi, Ota, Tanaka, Watanabe, and Maksyutov]{yokota2009global}
T~Yokota, Y~Yoshida, N~Eguchi, Y~Ota, T~Tanaka, H~Watanabe, and S~Maksyutov.
\newblock Global concentrations of co2 and ch4 retrieved from gosat: First preliminary results.
\newblock \emph{SOLA}, 5:\penalty0 160--163, 2009.
\newblock \doi{10.2151/sola.2009-041}.

\end{thebibliography}

\newpage
\section*{Appendix: Model Hyperparameters\label{appendix:A}}
This paper considers three models in addition to the baseline prediction. The baseline model only uses the xCO\textsubscript{2} measurements from OCO as a naive prediction for the ICOS and therefore it has no hyperparameters.

The Category Boosting model is based on the CatBoostClassifier model from the catboost package for python, using the following hyperparameters:
\begin{itemize}
    \item $nbr\_classes=25$
    \item $max\_depth=6$
    \item $learning\_rate=0.1$
    \item $iterations=100$
    \item $l2\_leaf\_reg=3$
\end{itemize}

The Extreme Gradient Boosting model is based on the XGBRegressor model from the XGBoost package for python, using the following hyperparameters:
\begin{itemize}
    \item $max\_depth=6$
    \item $learning\_rate=0.1$
    \item $n\_estimators=100$
    \item $gamma=0$
\end{itemize}

The Neural Network Regression model consists of 4 densely connected layers of sizes 64, 128, 64 and 32 respectively. 
%These use a Rectified Linear Unit (ReLU) activation function and l2 regularization. 
In addition, the model uses an output layer consisting of a single node with linear activation. The model uses a learning rate of 0.001.

In the internal layers of this neural network we employ a ReLU activation function which is one of the most widely used activation functions in neural networks, particularly in deep learning. It is defined as follows:
\[
\text{ReLU}(x) = \max(0, x) = \begin{cases} 
x & \text{if } x \geq 0 \\
0 & \text{if } x < 0 
\end{cases}
\]
Properties of ReLU:
\begin{itemize}
    \item Non-linearity: ReLU introduces non-linearity to the model, which is crucial for learning complex patterns.
    \item Computationally efficient: It is simple to compute, which speeds up the training process.
    \item Sparse activation: Since the ReLU outputs zero for all negative inputs, it often leads to sparse activations, making the network more efficient.
\end{itemize}
We also employ \( L_2 \) regularization which is a technique used to prevent overfitting in neural networks by penalizing large weights in the model. The basic idea is to add a penalty term to the loss function, which discourages the model from learning excessively large weights.

Given a loss function \( \mathcal{L}(\mathbf{y}, \hat{\mathbf{y}}) \), where \( \mathbf{y} \) are the true labels and \( \hat{\mathbf{y}} \) are the predicted labels, the \( L_2 \) regularized loss function is:

\[
\mathcal{L}_{\text{reg}}(\mathbf{y}, \hat{\mathbf{y}}) = \mathcal{L}(\mathbf{y}, \hat{\mathbf{y}}) + \lambda \sum_{i} w_i^2
\]

Here:
\begin{itemize}
    \item \( \mathcal{L}_{\text{reg}} \) is the regularized loss function.
    \item \( \mathcal{L} \) is the original loss function (which in our case refers to mean squared error).
    \item \( w_i \) represents the weights of the model.
    \item \( \lambda \) is the regularization parameter (often called the regularization strength), which controls the amount of regularization applied.
\end{itemize}

The \( L_2 \) regularization term \( \lambda \sum_{i} w_i^2 \) is the sum of the squares of the weights, multiplied by the regularization parameter \( \lambda \). This term encourages the model to keep the weights small, which helps to reduce overfitting.

\end{document}